\documentclass[10pt,twocolumn,letterpaper]{article}

\usepackage{cvpr}              

\usepackage[dvipsnames]{xcolor}

\definecolor{cvprblue}{rgb}{0.21,0.49,0.74}
\usepackage[pagebackref,breaklinks,colorlinks,citecolor=cvprblue,linkcolor=cvprblue]{hyperref}
\let\ul\underline
\usepackage{multirow}
\usepackage[normalem]{ulem}
\usepackage[table,xcdraw]{xcolor}
\useunder{\uline}{\ul}{}
\usepackage[T1]{fontenc}
\usepackage{colortbl}
\usepackage{adjustbox} 
\usepackage{amssymb, bbding}
\usepackage{amsmath}
\usepackage{bbding}
\usepackage{nicematrix}
\definecolor{colorTab}{rgb}{0.95,0.90,0.92}
\usepackage{graphicx}

\def\paperID{7870} 
\def\confName{CVPR}
\def\confYear{2026}

\title{Statistical Characteristic-Guided Denoising for \\ Rapid High-Resolution Transmission Electron Microscopy Imaging}

\author{
	Hesong Li\textsuperscript{1} 
	Ziqi Wu\textsuperscript{1} 
	Ruiwen Shao\textsuperscript{1}
	Ying Fu\textsuperscript{1$\dagger$}\\
	\vspace{5pt} 
	\textsuperscript{1}Beijing Institute of Technology\\
	\vspace{5pt} 
	{\tt\small \{lihesong2,wuziqi,rwshao,fuying\}@bit.edu.cn}
}
\usepackage{setspace}
\begin{document}
\maketitle
\renewcommand{\thefootnote}{}\footnote{$^\dagger$Corresponding author.}
\renewcommand{\thefootnote}{\arabic{footnote}}

\begin{abstract}
High-Resolution Transmission Electron Microscopy (HRTEM) enables atomic-scale observation of nucleation dynamics, which boosts the studies of advanced solid materials. Nonetheless, due to the millisecond-scale rapid change of nucleation, it requires short-exposure rapid imaging, leading to severe noise that obscures atomic positions. In this work, we propose a statistical characteristic-guided denoising network, which utilizes statistical characteristics to guide the denoising process in both spatial and frequency domains. In the spatial domain, we present spatial deviation-guided weighting to select appropriate convolution operations for each spatial position based on deviation characteristic. In the frequency domain, we present frequency band-guided weighting to enhance signals and suppress noise based on band characteristics. We also develop an HRTEM-specific noise calibration method and generate a dataset with disordered structures and realistic HRTEM image noises. It can ensure the denoising performance of models on real images for nucleation observation. Experiments on synthetic and real data show our method outperforms the state-of-the-art methods in HRTEM image denoising, with effectiveness in the localization downstream task. Code will
be available at \url{https://github.com/HeasonLee/SCGN}.
\end{abstract}

\section{Introduction}
The nucleation process of amorphous-to-crystalline transformation is a key mechanism in the formation of many solid materials. It has become an emerging research trend in advanced materials such as graphene \cite{graphene}, diamond \cite{diamond}, and metal catalysts \cite{UDVD}. The physical properties of these materials are closely related to the dynamic changes of atomic-scale structures. These dynamic processes usually occur on a millisecond time scale, thus requiring rapid imaging at the atomic level for observation. 

High-Resolution Transmission Electron Microscopy (HRTEM) imaging \cite{bf3} enables the capture of transient atomic dynamics during nucleation, providing an essential basis for understanding the formation mechanisms of advanced materials. However, short exposure times cause image signals to be submerged by various types of noise, making it difficult to identify atomic positions. Therefore, HRTEM image denoising is important for analyzing the nucleation process.

Traditional denoising methods, such as the Gaussian filter, cannot restore the correct atomic structure from strong HRTEM noise. With the development of deep learning, some existing methods \cite{AtomSegNet,SFIN} achieve denoising for crystal structures through convolutional neural networks. However, their denoising performance for the nucleation process is limited by networks and data. For network design, existing methods typically apply uniform processing to the overall features of the spatial or frequency domain. While in rapid imaging of disordered structures, they struggle to distinguish between signals and noise. For data, the ideal ground truth cannot be obtained through imaging, and there is a lack of denoising datasets for the nucleation process, which needs noise calibration specifically for HRTEM and simulation of disordered atomic arrangements. To summarize, a denoising network and a dataset specifically designed for nucleation observation are necessary.

In this work, we propose statistical characteristic-guided denoising for nucleation observation with rapid HRTEM imaging. Specifically, we first present statistical characteristic-guided weighting modules. Inspired by the traditional Yau-Yau filtering \cite{Yau}, which can effectively remove noise of non-linear systems via statistical characteristics of noise, our characteristic-guided weighting modules utilize statistical characteristics to guide the denoising process in both spatial and frequency domains. In the spatial domain, signals and noise at different positions exhibit distinct fluctuation characteristics, requiring different convolution operations for denoising. Thus, we introduce spatial deviation-guided weighting to select appropriate convolution operations for each spatial position based on the deviation characteristic. In the frequency domain, signals and noise are typically distributed in different frequency bands. Thus, we design frequency band-guided weighting to enhance signals and suppress noise based on band characteristics. Besides, we present an HRTEM-specific noise calibration method to generate a dataset with complex disordered structures and realistic noise for the observation of nucleation. It can boost our network's performance on real data. Experiments on synthetic and real data show our method outperforms the state-of-the-art methods in HRTEM image denoising, with effectiveness in the localization downstream task. 

In summary, our main contributions are that we

\begin{itemize}
\item Present spatial deviation-guided weighting, which can select appropriate convolutions for each spatial position based on the deviation characteristic;
\item Present frequency band-guided weighting, which can enhance signals and suppress noise in the frequency domain based on band characteristics;
\item Develop an HRTEM-specific noise calibration method to generate a dataset with complex disordered structures and realistic noise for observation of nucleation.
\end{itemize}

\section{Related Work}

\vspace{2mm}\noindent\textbf{Atomic image denoising methods.} 
Atomic image denoising methods are divided into two categories, \emph{i.e.}, traditional methods and deep learning methods. Traditional methods remove noise from images using traditional denoising methods such as the Gaussian filter, Wiener filter \cite{Wiener}, and bilateral filter \cite{Bilateral}. For the observation of nucleation, extremely low signal-to-noise ratios usually render traditional methods ineffective. Recently, deep learning is widely used in image processing \cite{cxh1,cxh2,gzj1,zt1,zt2,zyk1,zyk2,zyk3,gyp1,gyp2,gyp3,TriMSOD,CJE1,CJE2,CJE3,CJE4,hcm1,hcm2,3D-B2U}. For atomic image denoising, some methods \cite{FCN,AtomSegNet,UDVD,SFIN} learn the structural features of atomic arrangements from data through supervised or unsupervised learning. A fully convolutional network \cite{FCN} is used for denoising and identifying atomic deletions and replacements.  AtomSegNet \cite{AtomSegNet} is a convolutional neural network based on UNet architecture \cite{UNet} and plays a key role in studies on new battery materials \cite{Battery1,Battery2,Battery3}. UDVD \cite{UDVD} removes noise interference through unsupervised learning between multiple frames, but struggles to get clear atomic contours. SFIN \cite{SFIN} achieves better crystal atomic denoising results by incorporating frequency-domain processing into the network to utilize the strong frequency-domain features brought by the periodic arrangement of atoms. However, existing methods usually perform the same processing on all the features and cannot highlight atomic features from strong noise, resulting in limited denoising effects in observing nucleation.

\begin{figure}
    \centering
    \includegraphics[width=\linewidth]{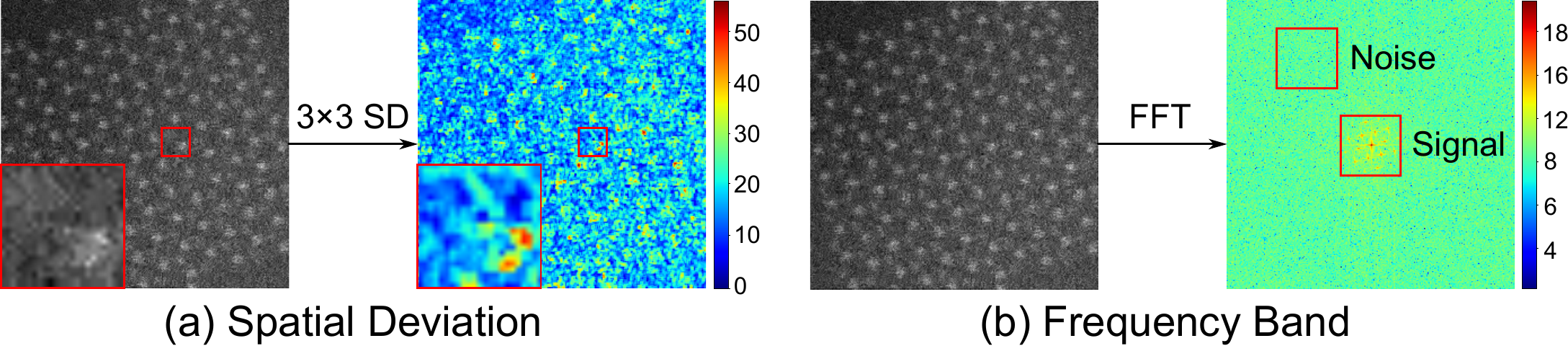}
    \vspace{-7mm}
    \caption{Visualization of the two kinds of statistical characteristics that we used to guide the denoising process. (a) The spatial deviation characteristic obtained by calculating the pixel standard deviation (SD) within $3\times3$ windows. It reflects the fluctuation conditions of different spatial positions in the image. (b) The frequency band characteristic obtained by the Fast Fourier Transform (FFT). Signal and noise are distributed at different frequency band positions and have different patterns.}
    \vspace{-3mm}
    \label{fig:moti}
\end{figure}

\vspace{2mm}\noindent\textbf{Atomic image denoising datasets.}
Due to the difficulty of obtaining ideal ground truths in actual experiments, synthetic datasets provide data support for deep learning-based denoising methods. The abTEM library \cite{abTEM} and the ReciPro software \cite{ReciPro} can simulate clean HRTEM images based on crystal structure files, but they struggle to simulate complex noises and the randomly disordered atomic arrangements during nucleation. The TEMImageNet dataset \cite{AtomSegNet} contains simulated noisy images and label images for transmission electron microscope image enhancement and atomic localization tasks. It includes several ordered atomic structures and incorporates scan noise, pointwise noise, and background with random parameters. SFIN dataset \cite{SFIN} simulates image noise using calibrated real parameters of Scanning Transmission Electron Microscopy (STEM) \cite{bf1, bf2}, which improves the reality of the simulated data and the performance of atom-level image denoising. However, different from STEM with pixel-by-pixel scanning, observation of nucleation requires HRTEM with more rapid global diffraction imaging. Therefore, HRTEM has different noise characteristics from STEM. In summary, there is currently a lack of noise calibration methods specifically for HRTEM, as well as datasets containing disordered structures for denoising of the nucleation process.

\section{Method}

In this section, we first formally define the HRTEM denoising problem and explain the motivation of our method. Then, we introduce our statistical characteristic-guided network with spatial deviation-guided weighting and frequency band-guided weighting. Finally, we introduce our data generation method for the observation of nucleation. 

\begin{figure*}
    \centering
    \includegraphics[width=.9\linewidth]{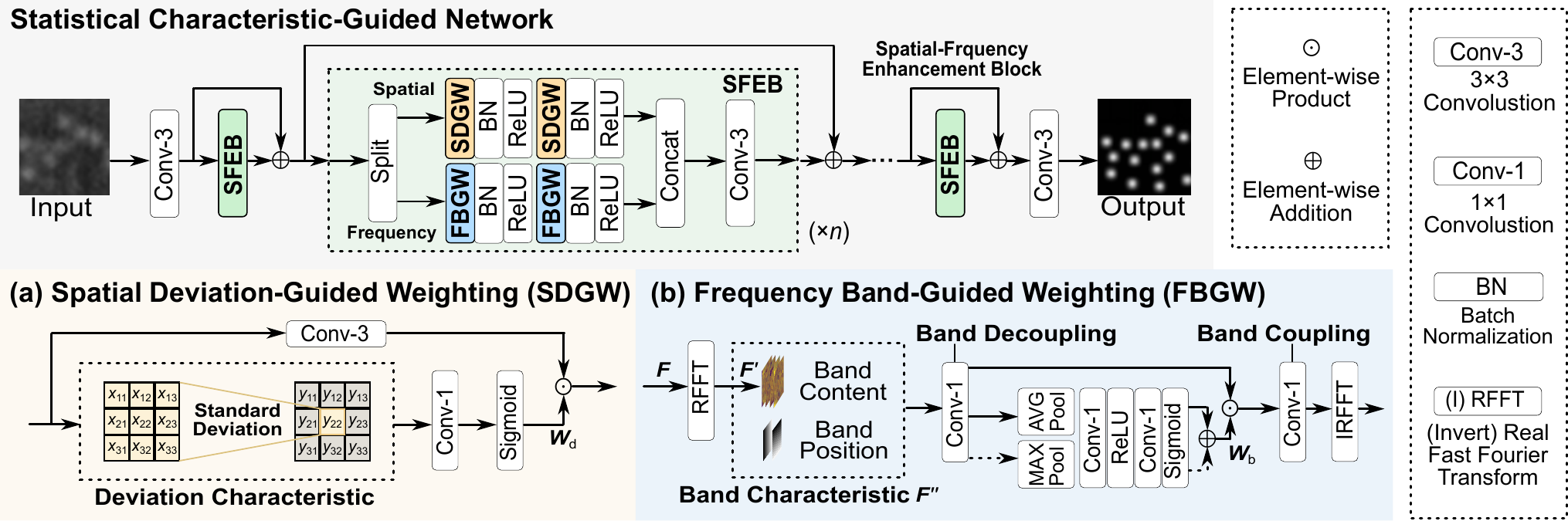}
    \vspace{-2mm}
    \caption{The framework of our statistical-characteristic guided network. It enhances signals and suppresses noise in the spatial and frequency domains using spatial deviation-guided weighting and frequency band-guided weighting, respectively.}
    \vspace{-2mm}
    \label{fig:frame}
\end{figure*}

\subsection{Formulation and Motivation}
The goal of HRTEM image denoising for the nucleation process is to restore the atomic structures from noisy HRTEM images, thereby analyzing the transition of atomic arrangement from disordered to ordered. Given a noisy image data $\boldsymbol{I}_\text{noisy}\in\mathbb{R}^{H \times W \times1}$, it is required to generate a clean image $\boldsymbol{I}_\text{clean}\in\mathbb{R}^{H \times W \times1}$, \emph{i.e.}, 
\begin{equation}
    \boldsymbol{I}_\text{clean}=\mathcal{G}(\boldsymbol{I}_\text{noisy})
\end{equation}
where $\mathcal{G}$ is the denoising method. The clean image $\boldsymbol{I}_\text{clean}$ has a black background, with a clear white dot at the location of each atom. High-speed imaging required for nucleation observation results in an extremely low signal-to-noise ratio of the input image $\boldsymbol{I}_\text{noisy}$. Existing networks \cite{AtomSegNet,SFIN} tend to mistake noise in the image for valid signals, leading to incorrect positions or missing atoms in the output clean image $\boldsymbol{I}_\text{clean}$. This challenge comes from two aspects, \emph{i.e.}, network and dataset. 

For \textbf{network}, in HRTEM images, intense noise is mixed with signals, making it difficult to distinguish signals from noise through ordinary convolution operations applied to the entire feature map. Yau-Yau filter \cite{Yau} is a traditional nonlinear filtering algorithm, whose filtering operations is specifically based on the statistical characteristics of signals and noise, effectively extracting signals from nonlinear systems. Inspired by this, we use statistical characteristics to guide the network in identifying signals and noise. Recent studies \cite{SFIN,FSNet,FDConv,FADC,FCDFusion} have proved the advantages of spatial and frequency domain processing and information fusion, especially for atom-level images \cite{SFIN}. Therefore, we consider describing signals and noise using statistical characteristics in both spatial and frequency domains. 
As shown in Figure~\ref{fig:moti}\textcolor{cvprblue}{(a)}, \textbf{spatial deviation} reflects the fluctuation conditions of different spatial positions in the image. Thus, spatial deviation can guide the network to perform different convolution operations on different spatial positions, thereby adapting to local regions with varying fluctuation characteristics. 
As shown in Figure~\ref{fig:moti}\textcolor{cvprblue}{(b)}, signals and noises are usually distributed in different \textbf{frequency bands}. Thus, frequency band positions and contents can guide the network to enhance signals and suppress noise. Based on above analysis, we compute dynamic weights for each component according to spatial deviation and frequency band, and guide the denoising process using these weights.

For \textbf{dataset}, existing datasets are not suitable for the denoising task of HRTEM images in the nucleation process.
\textbf{First}, the nucleation process requires observing the transformation of materials from an amorphous to a crystalline state, where most atomic arrangements are \textbf{disordered} or semi-disordered \cite{graphene,diamond,UDVD}. Deep learning techniques have recently been applied to atomic-scale image denoising, and existing methods primarily focus on denoising crystalline images. Thus, most synthetic datasets \cite{AtomSegNet} consist of fully ordered crystalline atomic structures, excluding the disordered structures required for nucleation. \textbf{Second}, the noise distribution of existing datasets is inconsistent with that of real HRTEM images. Studies on natural image denoising \cite{calib1,calib2} have proven that \textbf{noise calibration} helps synthesize more realistic noisy images and improve denoising performance. For atomic-scale images, most datasets \cite{GAN-data,AtomSegNet} randomly set noise parameters without calibration. Although SFIN dataset \cite{SFIN} calibrates STEM images, due to differences in imaging principles, the noise type and distribution of HRTEM are different from those of STEM. For example, HRTEM has column noise that STEM does not possess, while lacking the scan noise unique to STEM. Thus, direct use of existing datasets cannot guarantee noise adaptability. To address the above two issues, we generate disordered structures specific to the nucleation process and design a noise calibration method tailored for HRTEM.

\subsection{Overall Architecture}
The periodicity of atomic arrangements forms strong signals in the frequency domain, making it easy to distinguish overall signals from noise in the frequency domain. In contrast, the spatial domain facilitates the restoration of local features in different parts of the image.
Therefore, we design the spatial-frequency enhancement block, which leverages the advantages of both domains to enhance image features.
Specifically, as shown in Figure~\ref{fig:frame}, our statistical-characteristic guided network first extracts $C$-channel shallow features of a noisy HRTEM image $\boldsymbol{I}_\text{noisy}\in\mathbb{R}^{H \times W \times1}$ using a $3\times3$ convolution. It then uses $n$ spatial-frequency enhancement blocks with residual connections to gradually denoise and enhance the features. Finally, a $3\times3$ convolution fuses different features to generate a clean output image $\boldsymbol{I}_\text{clean}\in\mathbb{R}^{H \times W \times1}$.

Within each spatial-frequency enhancement block, the image features with $C$ channels are split equally into two parts, each with $C/2$ channels. One part is used for enhancement in the spatial domain, while the other is used for enhancement in the frequency domain.
In the spatial domain enhancement branch, we employ a convolution module with our spatial deviation-guided weighting to perform enhancement twice. In the frequency domain branch, we employ a convolution module with our frequency band-guided weighting to perform enhancement twice.
The enhanced results from the two branches are finally concatenated along the channel dimension, and a $3\times3$ convolution is applied to fuse features across different channels.

\subsection{Spatial Deviation-Guided Weighting}
We present a spatial deviation-guided weighting module, which can select appropriate convolution operations for each spatial position based on the deviation characteristic.

\vspace{2mm}\noindent\textbf{Deviation characteristic extraction.} 
Using different convolution operations for local regions with varying fluctuation characteristics can ensure the adaptability of denoising to noise conditions at different image positions. The standard deviation of pixels within a small window can reflect the local fluctuation in the image. Unlike linear statistics such as image gradients and means, the calculation of standard deviation involves nonlinear operations (\emph{e.g.}, computing the second moment includes squaring). Thus, it cannot be extracted through ordinary convolutions.
We design a window-based standard deviation module that can be embedded in convolutional networks. The local standard deviation of a single-channel feature map is defined as the standard deviation of pixel value distribution within a $3\times3$ window, \emph{i.e.},
\begin{equation}
    \text{SD}(\boldsymbol{X})=\sqrt{E[\boldsymbol{X}^2]-E[\boldsymbol{X}]^2+\epsilon}\label{eq}
\end{equation}
where the first moment $E[\boldsymbol{X}]$ is the mean of pixels in the window, which can be achieved via a $3\times3$ convolution with a fixed kernel. The second moment $E[\boldsymbol{X}^2]$ is obtained by squaring each element of $\boldsymbol{X}$ and then calculating the windowed mean. $\epsilon=10^{-5}$ is a small positive number introduced to ensure the expression under the square root remains positive during numerical calculations. For image edges, mirror padding is applied to add a one-pixel edge before the two $3\times3$ convolutions. This ensures the physical meaning of the standard deviation at edge positions remains unchanged. For each channel of the input feature map with $C$ channels, the $3\times3$ window standard deviation is calculated using Equation~\eqref{eq}, ultimately yielding a standard deviation feature map with $C$ channels.

\vspace{2mm}\noindent\textbf{Weighting process.} 
Using different convolution operations for different spatial positions is equivalent to applying a set of convolution operations to all spatial positions, followed by assigning different weights to the results of these operations. As shown in Figure~\ref{fig:frame}\textcolor{cvprblue}{(a)}, our spatial deviation-guided weighting consists of two branches.
The upper branch extracts features using a $3\times3$ convolution, where different channels represent distinct convolution operations. The lower branch generates the dynamic weight $\boldsymbol{W}_\text{d}\in\mathbb{R}^{H \times W \times C/2}$ based on the extracted standard deviation feature map through a $1\times1$ convolution and a sigmoid function. Finally, the feature map from the upper branch is weighted via element-wise multiplication.
In the lower branch, the $1\times1$ convolution integrates standard deviation features across different channels to determine weights. The sigmoid function normalizes the weight values to the range $[0, 1]$. For each spatial position, the $1\times1$ convolution and sigmoid function play as a one-dimensional neural network composed of a fully connected layer and an activation layer, which assigns specific weights according to the fluctuation characteristics of the position.

\subsection{Frequency Band-Guided Weighting}
We present a frequency band-guided weighting module, which can enhance signals and suppress noise in the frequency domain based on band characteristics.

\vspace{2mm}\noindent\textbf{Band characteristic extraction.} 
Due to periodic differences, signals and noise are typically distributed in different frequency bands in the frequency domain. To enhance signals and suppress noise, it is necessary to determine whether a frequency band is a signal or noise based on its position and content, and assign corresponding weights accordingly.
To obtain the content of frequency bands, we convert the image feature $\boldsymbol{F}$ to the frequency domain using the Real Fast Fourier Transform (RFFT) in Pytorch \cite{PyTorch}. Since the result is a complex tensor $\boldsymbol{F}'$ containing imaginary and real parts, the number of channels becomes $2C$.
As the feature map $\boldsymbol{F}'$ lacks position coordinate information, we explicitly add a position embedding feature map with two channels, corresponding to the relative coordinate values in the horizontal and vertical directions, respectively, with values ranging from $0$ to $1$. This yields a frequency-domain feature $\boldsymbol{F}''$ with $2C+2$ channels, which includes both band content and band position information.
 
\vspace{2mm}\noindent\textbf{Weight computation.} 
As shown in Figure~\ref{fig:frame}\textcolor{cvprblue}{(b)}, after obtaining the band characteristic feature $\boldsymbol{F}''$, a $1\times1$ convolution is used to decouple frequency bands. This decoupling process assigns frequency band content of different features to distinct feature channels—some channels represent valid signals, while others correspond to noise irrelevant to the signals.
Subsequently, a frequency band classification module identifies the importance of each channel based on the features of different channels and assigns appropriate weights. We design this classification module with reference to the channel attention mechanism \cite{CA}, which sequentially includes global pooling, $1\times1$ convolution, ReLU activation, $1\times1$ convolution, and a sigmoid function. Among these components, global pooling is used to convert the global information of each channel into a scalar.
We employ average pooling and max pooling, respectively, as the global pooling operations, and sum the results of the two corresponding classification modules. Thus, each channel of the weight $\boldsymbol{W}_\text{b}\in\mathbb{R}^{1 \times 1 \times C/2}$ is in the range of $[0, 2]$. Weights greater than $1$ correspond to enhancement, while those less than $1$ correspond to suppression.

\vspace{2mm}\noindent\textbf{Weighting process.} 
As shown in Figure~\ref{fig:frame}\textcolor{cvprblue}{(b)}, the weighting process involves multiplying each entire channel of $\boldsymbol{F}''$ by its corresponding weight, thereby obtaining each frequency band after enhancement or suppression. These frequency bands are subsequently recoupled into a complete frequency-domain feature using a $1\times1$ convolution. Finally, an inverse real fast Fourier transform from Pytorch \cite{PyTorch} is applied to convert the frequency-domain feature back to the spatial domain.

\begin{figure}
    \centering
    \includegraphics[width=\linewidth]{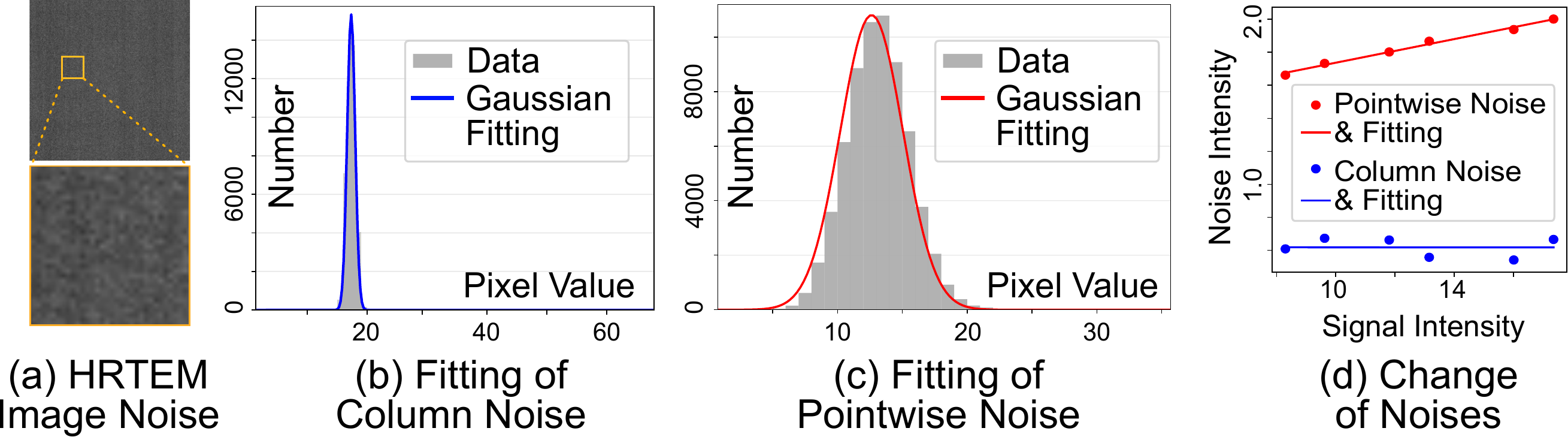}
    \vspace{-7mm}
    \caption{HRTEM image noises and their calibration. Column noise and pointwise noise can be fitted with a Gaussian distribution. The intensity of pointwise noise is proportional to the signal intensity. The intensity of column noise is almost a constant value.}
    \label{fig:calib}
\end{figure}

\subsection{Data Generation}
To generate an HRTEM dataset for observing nucleation, we first generate clean images and then add noise to them.

\vspace{2mm}\noindent\textbf{Clean image generation.} 
Most atoms in the nucleation process are arranged in a disordered manner, but due to interatomic forces, even the two closest atoms maintain a certain distance. Directly using random dot placement would cause atomic overlaps, which is inconsistent with real scenarios. To address this issue, we use the Poisson disk sampling \cite{Poisson} to generate the positions of each atom. This algorithm can generate a two-dimensionally distributed point set with linear complexity, ensuring that the distance between any two points is greater than a threshold (\emph{e.g.}, 4 pixels) while maintaining an overall random arrangement of all points. Then, binarized 2D Perlin noise \cite{Perlin} is used as a mask to remove some atomic positions, simulating the vacuum region outside the material.

After obtaining the position of each atom, each atom is rendered onto the image as a 2D Gaussian distribution. The brightness and size of the atoms (\emph{i.e.}, the peak brightness and standard deviation of the 2D Gaussian distribution), as well as the background brightness of the image, are all derived from measurements of real HRTEM images, following the SFIN dataset \cite{SFIN}. When generating clean HRTEM images, the brightness and size of atoms exhibit certain fluctuations to simulate aberrations in the imaging process and blurring caused by multiple scattering. In contrast, when generating ground truth maps, the atomic brightness is fixed at 255, the standard deviation of the normal distribution is fixed at 0.75, and the image background brightness is fixed at 0. This ensures that the ground truth is clear.

\vspace{2mm}\noindent\textbf{HRTEM noise calibration and generation.} 
Most HRTEMs convert the electron beam transmitted through the sample into an optical signal using a photoelectric converter and then receive the optical signal with a CMOS sensor. The image noise mainly originates from the CMOS sensor, including dot noise $\boldsymbol{N}_\text{p}$ (with independent values for each pixel) and column noise $\boldsymbol{N}_c$ (with identical values for all pixels in a vertical column). The noisy HRTEM image can be formulated as
\begin{equation}
\boldsymbol{I}_\text{noisy}=\boldsymbol{I}_\text{clean}+\boldsymbol{N}_\text{c}+\boldsymbol{N}_\text{p}(\boldsymbol{I}_\text{clean})
\end{equation}
where $\boldsymbol{I}_\text{noisy}$ and $\boldsymbol{I}_\text{clean}$ are the noisy image and the clean image, respectively. $\boldsymbol{N}_\text{p}$ is the pointwise noise related to the signal intensity $\boldsymbol{I}_\text{clean}$. 

To calibrate column noise and pointwise noise, we capture six sequences of images of the vacuum region using HRTEM. These images contain only noise without any atoms, and one example is shown in Figure~\ref{fig:calib}\textcolor{cvprblue}{(a)}. We read the unadjusted RAW image data from the Electron Microscopy Dataset (EMD) files saved by the Velox software \cite{Velox}. Each sequence is captured under a different electron beam intensity (\emph{i.e.}, signal intensity), ranging from 1 nA to 2 nA with an interval of 0.2 nA. For each sequence, 100 frames are captured to fit the noise parameters. As shown in Figure~\ref{fig:calib}\textcolor{cvprblue}{(b)} and Figure~\ref{fig:calib}\textcolor{cvprblue}{(c)}, both the pointwise noise and column noise in a single image can be well-fitted by a normal distribution. Figure~\ref{fig:calib}\textcolor{cvprblue}{(d)} shows that the intensity of pointwise noise is proportional to the signal intensity, while the intensity of column noise is almost a constant value. Based on the data from all six sequences, the relationship between the amplitude of pointwise noise and signal intensity is obtained as
\begin{equation}
\sigma_\text{p}(x,y)=0.03583 \boldsymbol{I}_\text{clean}(x,y)+1.379
\end{equation}
where $\sigma_\text{p}(x,y)$ and $\boldsymbol{I}_\text{clean}(x,y)$ are the standard deviation of pointwise noise $\boldsymbol{N}_\text{p}$ and the pixel value of the clean image $\boldsymbol{I}_\text{clean}$ at the location $(x,y)$, respectively. The standard deviation of column noise $\boldsymbol{N}_\text{c}$ is a constant value $\sigma_\text{c}=0.6641$. Gaussian noise is added to each pixel in the clean image based on the calibrated noise parameters $\sigma_\text{p}$ and $\sigma_\text{c}$ to form a noisy HRTEM image.

\section{Experiments}
In this section, we first introduce the experimental settings. Then, we provide the main results of different methods on HRTEM denoising, as well as their application results on atomic localization. Finally, we discuss the effectiveness of our network and dataset.  

\subsection{Experimental Settings}
We first introduce the datasets and metrics. Then, we introduce the comparison methods and implementation details of our network.

\vspace{2mm}\noindent\textbf{Datasets and metrics.}
We follow SFIN \cite{SFIN} to compare different methods on both synthetic datasets and real data. The synthetic datasets are TEMImageNet \cite{AtomSegNet}, SFIN dataset \cite{SFIN}, and our NUC dataset for \textbf{nuc}leation. The real data is a sequence of frames actually captured by HRTEM, which includes the nucleation process of Pt (platinum) transitioning from disordered to ordered. We follow SFIN \cite{SFIN} to use Peak Signal-to-Noise Ratio (PSNR) and Structural Similarity Index Measure (SSIM) \cite{SSIM} as metrics.

\vspace{2mm}\noindent\textbf{Comparison methods.}
We compare our method with the Gaussian filter, AtomSegNet \cite{AtomSegNet}, SFIN \cite{SFIN}, HINT \cite{HINT}, and UDVD \cite{UDVD}. The Gaussian filter is a traditional method. AtomSegNet, SFIN, and HINT are supervised deep learning methods. UDVD is an unsupervised deep learning method. Since UDVD requires multi-frame data as input, comparisons with it are only conducted using multi-frame real data.

\vspace{2mm}\noindent\textbf{Implementation details of our network.} 
Our statistical characteristic-guided network is trained with Adam optimizer \cite{Adam} ($\beta_1$ = 0.9 and $\beta_2$ = 0.999) and $L_1$ loss for 100 epochs on our dataset NUC (with 1000 pairs of training data). The block number $n=8$, and the channel number $C=64$. The codes are based on PyTorch \cite{PyTorch}. The learning rate is set as $2\times10^{-4}$. Mini-batch size is set to 6, which ensures that training can finish on a single RTX 3090 GPU.

\subsection{Main Results}

We compare our Statistical Characteristic-Guided Network (SCGN) with the state-of-the-art methods on both synthetic datasets and real data.

\begin{figure}
    \centering
    \includegraphics[width=\linewidth]{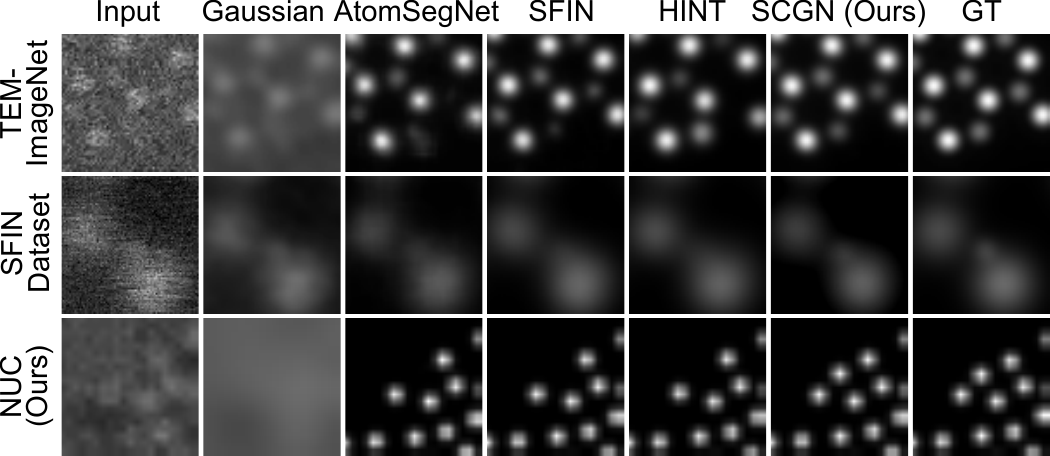}
    \vspace{-7mm}
    \caption{Denoising results on different datasets. Training datasets correspond to the testing datasets.}
    \label{fig:denoising}
    \vspace{-2mm}
\end{figure}

\vspace{2mm}\noindent\textbf{Results on synthetic datasets.} Figure~\ref{fig:denoising} shows the results on different datasets, including TEMImageNet \cite{AtomSegNet}, SFIN dataset \cite{SFIN}, and our NUC dataset. In the results on TEMImageNet, the result of Gaussian filter is blurry, while in the results of AtomSegNet \cite{AtomSegNet}, SFIN \cite{SFIN}, and HINT \cite{HINT}, some darker atoms cannot be recovered. Our SCGN can clearly display both brighter and darker atoms. In the results on SFIN dataset, only our SCGN clearly distinguishes the two atoms in the lower right part of the image. In the results on our NUC dataset, the result of Gaussian filter is blurry, while in the results of AtomSegNet, SFIN, and HINT, an atom is missed. Our SCGN can clearly display all the atoms. Table~\ref{tab:main} shows the quantitative results. Benefiting from the weighting of signals and noise, our SCGN outperforms other methods on both existing datasets and our dataset. On our NUC dataset, the PSNR exceeds that of the best-performing SFIN method by 0.85 dB.

   \begin{table}[]\small
\centering
\caption{Denoising results of different methods on different datasets. Training datasets correspond to the testing datasets. Best and second best values are \textbf{bold} and {\ul underlined}, respectively. Our SCGN achieves the best denoising results.}
\vspace{-2mm}
\setlength{\tabcolsep}{.5mm}
\resizebox{1\columnwidth}{!}{
\begin{NiceTabular}{c|cc|cc|cc}
\hline
                                                                            & \multicolumn{2}{c|}{TEMImageNet \cite{AtomSegNet}} & \multicolumn{2}{c|}{SFIN Dataset \cite{SFIN}} & \multicolumn{2}{c}{NUC (Ours)} \\
\multirow{-2}{*}{Method}                                                    & PSNR $\uparrow$          & SSIM $\uparrow$         & PSNR $\uparrow$       & SSIM $\uparrow$       & PSNR $\uparrow$ & SSIM $\uparrow$ \\ \hline
Gaussian                                                                    & 16.08                    & 0.4991                  & 16.43                 & 0.3267                & 8.26            & 0.0207          \\
AtomSegNet \cite{AtomSegNet} \textcolor{gray}{\footnotesize [Sci. Rep.'22]} & 22.34                    & 0.8008                  & 34.33                 & 0.8994                & 23.44           & 0.9452          \\
SFIN \cite{SFIN} \textcolor{gray}{\footnotesize [CVPR'25]}                  & {\ul 31.76}              & {\ul 0.9543}            & {\ul 38.48}           & {\ul 0.9582}          & {\ul 25.70}     & {\ul 0.9673}    \\
HINT \cite{HINT} \textcolor{gray}{\footnotesize [ICCV'25]}                  & 27.51                    & 0.9031                  & 37.62                 & 0.9541                & 23.53           & 0.9441          \\
\rowcolor[HTML]{F2F2F2} 
SCGN (Ours)                                                                 & \textbf{32.55}           & \textbf{0.9638}         & \textbf{38.99}        & \textbf{0.9634}       & \textbf{26.55}  & \textbf{0.9717} \\ \hline
\end{NiceTabular}}\label{tab:main}
 \end{table}

 \begin{figure*}
    \centering
    \includegraphics[width=.9\linewidth]{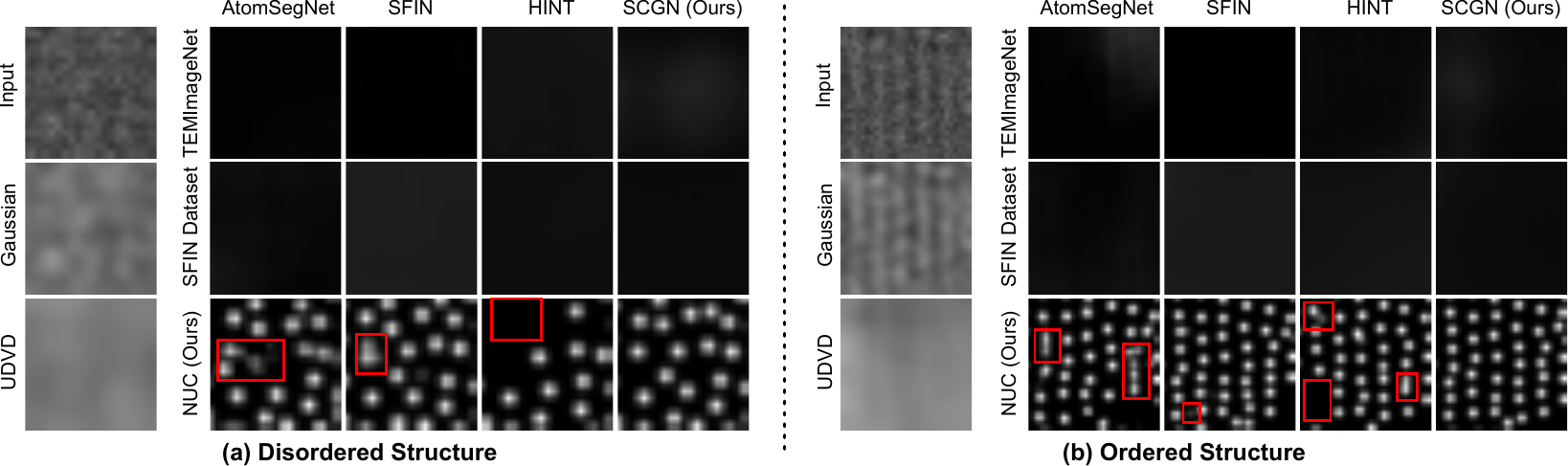}
     \vspace{-3mm}
    \caption{Denoising results on real data. Gaussian filter is a traditional method. UDVD \cite{UDVD} is an unsupervised learning method. The right parts of (a) and (b) are the results of supervised learning methods trained on different datasets. Red boxes mark some errors.}
    \label{fig:denoising-real}
    \vspace{-5mm}
\end{figure*}

\vspace{2mm}\noindent\textbf{Results on real data.} 
The denoising results on real data are shown in Figure~\ref{fig:denoising-real}. It shows the effects of different methods and different training datasets. Although the result of Gaussian filtering is free of noise, it is difficult to distinguish between different atoms. The results of UDVD are even blurrier than those of the Gaussian filter. The reason is that the content varies significantly between frames during the nucleation process, making it difficult to perform mutual supervised learning through adjacent frames. The right parts of Figure~\ref{fig:denoising-real}\textcolor{cvprblue}{(a)} and Figure~\ref{fig:denoising-real}\textcolor{cvprblue}{(b)} show the performance of different supervised learning methods trained on different datasets. Results trained using existing datasets can barely identify atoms. The reason is that the noise distribution of existing datasets differs from that of HRTEM images, and they lack densely and disorderly arranged atoms in the nucleation process. As shown in the last row, our noise calibration method and disordered atom generation method enable most networks to recover atomic structures from real HRTEM images. As marked by the red boxes, the denoising results of the AtomSegNet \cite{AtomSegNet}, SFIN \cite{SFIN}, and HINT \cite{HINT} contain various errors, such as missing atoms, extra atoms, and adhesion of multiple atoms. Our SCGN has no such errors when denoising both disordered (Figure~\ref{fig:denoising-real}\textcolor{cvprblue}{(a)}) and ordered (Figure~\ref{fig:denoising-real}\textcolor{cvprblue}{(b)}) atomic arrangements, which indicates that our method facilitates the accurate observation of the nucleation process.

\subsection{Application on Atomic Localization}
After obtaining the denoising results, the position of each atom can be easily located. It plays a key role in the automated analysis of the structural properties of materials \cite{AtomAI,AtomSegNet}. We follow AtomSegNet \cite{AtomSegNet} and SFIN \cite{SFIN} to compare the atomic localization results of different methods, and use Intersection over Union (IoU) metric for quantitative evaluation. 

\begin{figure}
    \centering
    \includegraphics[width=.9\linewidth]{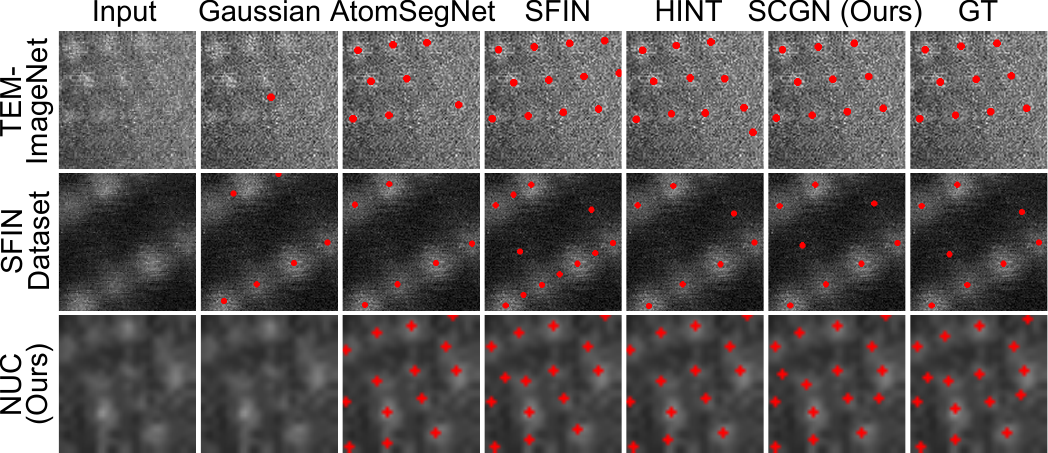}
    \vspace{-3mm}
    \caption{Localization results on different datasets. Training datasets correspond to the testing datasets.}
    \label{fig:local}
    \vspace{-2mm}
\end{figure}

\begin{figure}
    \centering
    \includegraphics[width=.9\linewidth]{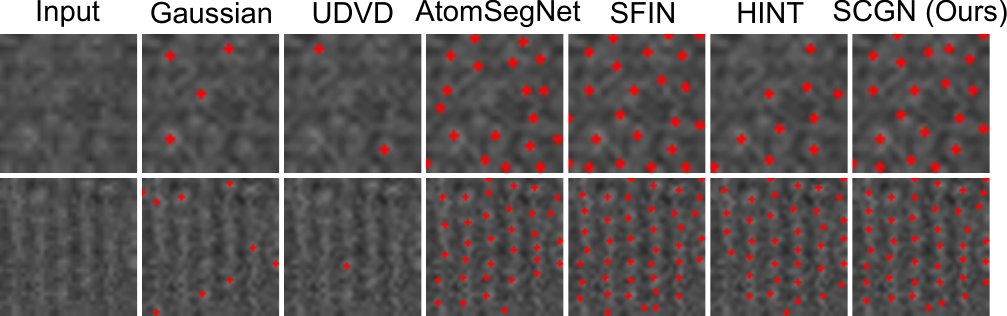}
     \vspace{-3mm}
    \caption{Localization results on real data. Supervised learning methods are trained on our NUC dataset.}
    \label{fig:local-real}
    \vspace{-2mm}
\end{figure}

Figures~\ref{fig:local} and \ref{fig:local-real} show the localization results on three synthetic datasets and real data, respectively. It can be observed that the results of Gaussian filtering, UDVD \cite{UDVD}, AtomSegNet \cite{AtomSegNet}, SFIN \cite{SFIN}, and HINT \cite{HINT} contain false detections and missed detections. The reason is that the signal and noise were not well distinguished in the denoising stage. Our SCGN distinguishes signal and noise via statistical characteristic-guided weighting and achieves a better atomic localization performance. Table~\ref{tab:local} also shows that our SCGN outperforms other methods on the three datasets.

   \begin{table}[]\small
\centering
\caption{Localization results (IoU metrics) of different methods on different datasets. Best and second best values are \textbf{bold} and {\ul underlined}, respectively.}
\vspace{-2mm}
\resizebox{1\columnwidth}{!}{
\setlength{\tabcolsep}{.5mm}
\begin{NiceTabular}{c|ccc}
\hline
Method                                                                      & TEMImageNet \cite{AtomSegNet} & SFIN Dataset \cite{SFIN} & NUC (Ours)   \\ \hline
Gaussian                                                                    & 0.1416                        & 0.0736                   & 0.0940          \\
AtomSegNet \cite{AtomSegNet} \textcolor{gray}{\footnotesize [Sci. Rep.'22]} & 0.5636                        & 0.4507                   & 0.6393          \\
SFIN \cite{SFIN} \textcolor{gray}{\footnotesize [CVPR'25]}                  & {\ul 0.7429}                  & {\ul 0.5694}             & {\ul 0.7124}    \\
HINT \cite{HINT} \textcolor{gray}{\footnotesize [ICCV'25]}                  & 0.6360                        & 0.4924                   & 0.6494          \\
\rowcolor[HTML]{F2F2F2} 
SCGN (Ours)                                                                 & \textbf{0.7444}               & \textbf{0.6083}          & \textbf{0.7317} \\ \hline
\end{NiceTabular}}\label{tab:local}
 \end{table}

\subsection{Discussion}
We discuss the effectiveness of our spatial deviation-guided weighting, our frequency band-guided weighting, and our NUC dataset in order.

  \begin{table}[]\small
\centering
\caption{Ablation study on our Spatial Deviation-Guided Weighting (SDGW) and Frequency Band-Guided Weighting (FBGW). ``PE'' means position embedding in FBGW. ``CA'' means using channel attention \cite{CA} instead of FBGW. Models are tested on our NUC dataset.}
\vspace{-2mm}
\resizebox{.85\columnwidth}{!}{
\begin{NiceTabular}{c|ccc|cc}
\hline
Method      & SDGW         & FBGW         & PE           & PSNR $\uparrow$ & SSIM $\uparrow$ \\ \hline
SCGN-V1     & \XSolidBrush & \XSolidBrush & \XSolidBrush & 25.74           & 0.9672          \\
SCGN-V2     & \XSolidBrush & \Checkmark   & \Checkmark   & 26.23           & 0.9699          \\
SCGN-V3     & \Checkmark   & \XSolidBrush & \XSolidBrush & 26.19           & 0.9681          \\
SCGN-V4     & \Checkmark   & CA           & \XSolidBrush & 26.24           & 0.9689          \\
SCGN-V5     & \Checkmark   & \Checkmark   & \XSolidBrush & 26.48           & 0.9710          \\
\rowcolor[HTML]{F2F2F2} 
SCGN (Ours) & \Checkmark   & \Checkmark   & \Checkmark   & \textbf{26.55}  & \textbf{0.9717} \\ \hline
\end{NiceTabular}}\label{tab:ablation}
 \end{table}
 
\begin{figure}
    \centering
    \includegraphics[width=.9\linewidth]{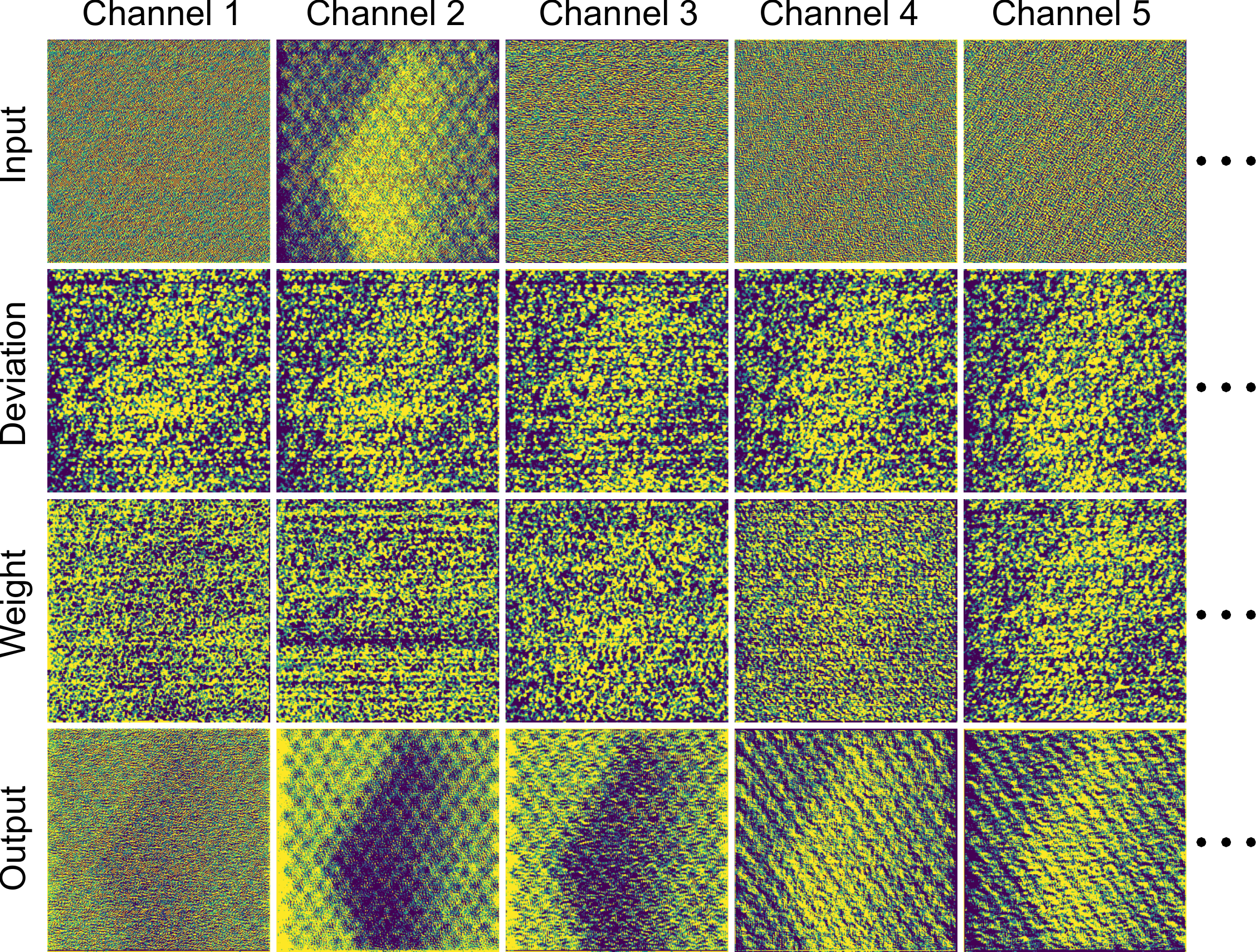}
    \vspace{-2mm}
    \caption{Visualization of our spatial deviation-guided weighting. By generating weights ($\boldsymbol{W}_\text{d}$ in Figure~\ref{fig:frame}\textcolor{cvprblue}{(a)}) for each spatial position based on local standard deviation, our spatial deviation-guided weighting can dynamically adjust the contribution of different convolution operations to different positions, thereby restoring atomic structures in complex noises.}
    \label{fig:vis_std}
     \vspace{-2mm}
\end{figure}

\vspace{2mm}\noindent\textbf{Effect of spatial deviation-guided weighting.}
Our spatial deviation-guided weighting generates weights for each spatial position based on local standard deviation. It can dynamically adjust the contribution of different convolution operations to different positions. Figure~\ref{fig:vis_std} visualizes the input features, standard deviation maps, weight maps, and output features. The weights in the same channel vary across different spatial positions, indicating that the weighting operation is pixel-wise. Compared with the input feature map, which is almost entirely composed of noise, the feature map obtained through dynamically weighted convolution operations contains more distinct structural information. Results in Table~\ref{tab:ablation} show that our spatial deviation-guided weighting improves PSNR by 0.32 dB (SCGN-V2$\rightarrow$SCGN) to 0.45 dB (SCGN-V1$\rightarrow$SCGN-V3).

\begin{figure}
    \centering
    \includegraphics[width=.9\linewidth]{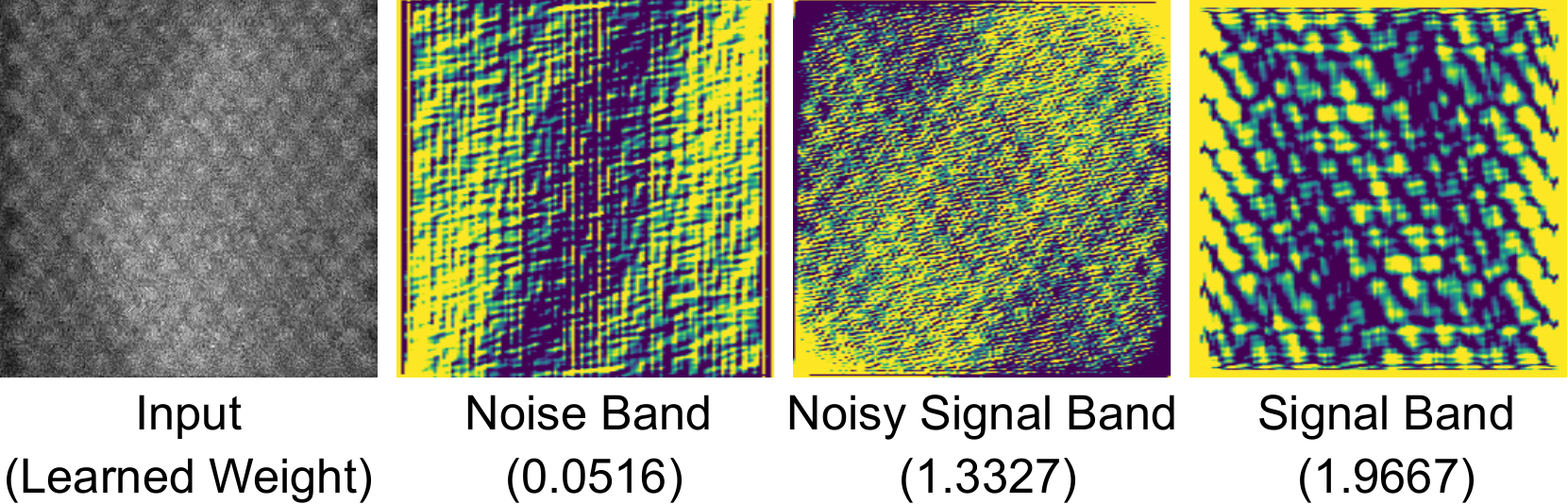}
    \vspace{-2mm}
    \caption{Visualization of our frequency band-guided weighting. Three representative frequency bands from the third frequency band-guided weighting module are converted to the spatial domain and visualized. It shows that our frequency band-guided weighting can highlight signals and suppress noise by giving different weights ($\boldsymbol{W}_\text{b}$ in Figure~\ref{fig:frame}\textcolor{cvprblue}{(b)}) to them.}
    \label{fig:vis_freq}
    \vspace{-2mm}
\end{figure}

\vspace{2mm}\noindent\textbf{Effect of frequency band-guided weighting.}
Our frequency band-guided weighting enhances signals and suppresses noise by accounting for differences in frequency band content and position. Figure~\ref{fig:vis_freq} shows examples of weighting applied to three different frequency bands.
First, the feature map is decoupled into multiple frequency bands in the frequency domain. For noise bands irrelevant to the signal, the weights learned by the network are close to 0, and these bands are suppressed. For noisy signal bands that are related to the signal but have unclear structures, the learned weights are close to 1, so these bands are retained. For signal bands that are related to the signal and have clear structures, the learned weights are close to 2, and these bands are enhanced.
After weighting, the signal is enhanced and the noise is reduced.

Unlike channel attention \cite{CA}, which weights different feature channels in the spatial domain, our frequency band-guided weighting operates in the latent space of the frequency domain. This makes it easier to distinguish periodic signals of atomic arrangements from spatially independent noise by leveraging the frequency band distribution characteristics of signals and noise.
The comparison results in Table~\ref{tab:ablation} show that using frequency band-guided dynamic weighting can improve PSNR by 0.36 dB (SCGN-V3 $\rightarrow$ SCGN), outperforming channel attention (SCGN-V4). Furthermore, using both frequency band content and position information (SCGN) yields better results than using only content information (SCGN-V5). This indicates that the position embedding is necessary.

\vspace{2mm}\noindent\textbf{Effect of our dataset.}
Table~\ref{tab:data} compares our NUC dataset with existing atomic-level transmission electron microscopy datasets, \emph{i.e.}, TEMImageNet \cite{AtomSegNet} and SFIN Dataset \cite{SFIN}. Our NUC dataset is more suitable for observing nucleation in both clean image generation and noise generation.
It has non-overlapping and disordered atomic arrangements, and uses HRTEM noise calibration considering column noise. Figure~\ref{fig:denoising-real} shows that using our dataset can denoise real nucleation images, which cannot be easily achieved using other datasets.

   \begin{table}[]\small
\centering
\caption{Comparison of different datasets.}
\vspace{-3mm}
\resizebox{.9\columnwidth}{!}{
\setlength{\tabcolsep}{.5mm}
\begin{NiceTabular}{c|cc|cc}
\hline
                              & \multicolumn{2}{c|}{Clean Image Generation} & \multicolumn{2}{c}{Noise Generation} \\ \cline{2-5} 
                              & Non-                 & Disordered           & HRTEM             & Column           \\
\multirow{-3}{*}{Dataset}     & Overlapping          & Arrangement          & Calibration       & Noise            \\ \hline
TEMImageNet \cite{AtomSegNet} & \Checkmark           & \XSolidBrush         & \XSolidBrush      & \XSolidBrush     \\
SFIN Dataset \cite{SFIN}      & \XSolidBrush         & \Checkmark           & \XSolidBrush      & \XSolidBrush     \\
\rowcolor[HTML]{F2F2F2} 
NUC (Ours)                    & \Checkmark           & \Checkmark           & \Checkmark        & \Checkmark       \\ \hline
\end{NiceTabular}}\label{tab:data}
 \end{table}
 \vspace{2mm}
\section{Conclusion}
In this paper, inspired by the Yau-Yau filter, we propose statistical characteristic-guided denoising for HRTEM images to enable clear observation of the nucleation process.
First, our spatial deviation-guided weighting module dynamically adjusts convolution operations based on the standard deviation of local features, allowing it to adapt to signal and noise conditions at different positions.
Second, our frequency band-guided weighting module distinguishes signals from noise in the frequency domain based on the position and content of frequency bands, which can enhance signals and suppress noise.
Finally, we establish the NUC dataset tailored to the nucleation process, which includes more realistic disordered atomic arrangements and HRTEM image noise.
Our method achieves high-precision atomic image denoising on three synthetic datasets and real data, facilitating the accurate localization of atomic positions.
In the future, to design a more general filter for various noises in different kinds of images, it is worth investigating more statistical characteristics to guide weighting, such as the median and higher-order moments of local image regions.

\section*{Acknowledgements}
This work is supported by the National Natural Science Foundation of China (62331006), and the Fundamental Research Funds for the Central Universities.

{
    \small
    \bibliographystyle{ieeenat_fullname}
    \bibliography{main}
}


\end{document}